\documentclass[table,dvipsnames,svgnames]{article}
\usepackage{aaai2026}  
\usepackage{times}  
\usepackage{helvet}  
\usepackage{courier}  
\usepackage[hyphens]{url}  
\usepackage{graphicx} 
\usepackage{natbib}  
\usepackage{caption} 
\usepackage{algorithm}
\usepackage{algorithmic}
\usepackage{booktabs}
\usepackage{newfloat}
\usepackage{listings}
\DeclareCaptionStyle{ruled}{labelfont=normalfont,labelsep=colon,strut=off} 
\floatstyle{ruled}
\newfloat{listing}{tb}{lst}{}
\floatname{listing}{Listing}
\usepackage{hyperref}
\nocopyright

\usepackage{subcaption} 
\newcommand{\our}{{\textsc{TrajEvo}}}
\usepackage{amsmath}
\usepackage{amssymb}
\usepackage[capitalise,nameinlink]{cleveref}
\usepackage{graphicx}
\usepackage{booktabs}
\usepackage{multirow}
\usepackage{array}

\lstdefinestyle{promptstyle_main}{
    basicstyle=\scriptsize\ttfamily,
    keywordstyle=\bfseries,           
    commentstyle=\itshape,           
    frame=tb,                       
    framesep=4pt,                   
    breakautoindent=false,
    breakindent=0pt, 
    breakatwhitespace=true,
    breaklines=true,
    captionpos=b,
    columns=flexible,
    extendedchars=true,
    fontadjust=true,
    inputencoding=utf8,
    keepspaces=true,
    lineskip=-0.1pt,
    linewidth=0.98\columnwidth,
    resetmargins=true,
    showspaces=false,
    showstringspaces=false,
    showtabs=false,
    tabsize=2,
    upquote=true,
    moredelim=[s]{\{}{\}},
    morekeywords={Index, Count}, 
    morecomment=[l]{<},            
    numbers=none,
    xleftmargin=8pt,
    xrightmargin=8pt,
    aboveskip=0.8\baselineskip,
    belowskip=0.8\baselineskip
}

\lstdefinestyle{promptstyle_appendix}{
    backgroundcolor=\color{gray!8},
    basicstyle=\scriptsize\ttfamily\color{black!90},
    breakautoindent=false,
    breakindent=0pt, 
    breakatwhitespace=true,
    breaklines=true,
    captionpos=b,
    keepspaces=true,
    showspaces=false,
    showstringspaces=false,
    showtabs=false,
    frame=single,
    rulecolor=\color{gray!40},
    framesep=3pt,
    frameround=tttt,
    framexleftmargin=6pt,
    xleftmargin=8pt,
    xrightmargin=8pt,
    tabsize=2,
    linewidth=0.98\textwidth,
    fontadjust=true,
    numbers=none,
    aboveskip=0.8\baselineskip,
    belowskip=0.8\baselineskip,
    columns=flexible,
    upquote=true,
    inputencoding=utf8,
    extendedchars=true,
    lineskip=-0.1pt,
    resetmargins=true,
    moredelim=[s][\color{blue!70!black}]{\{}{\}},
}

\lstdefinestyle{heuristicstyle}{
    backgroundcolor=\color{gray!5},
    commentstyle=\color{green!60!black},
    keywordstyle=\color{blue!70!black}\bfseries,
    numberstyle=\tiny\color{gray!70},
    stringstyle=\color{purple!70!black},
    basicstyle=\scriptsize\ttfamily\color{black},
    breakatwhitespace=false,
    breaklines=true,
    captionpos=b,
    keepspaces=true,
    showspaces=false,
    showstringspaces=false,
    showtabs=false,
    tabsize=4,
    frame=single,
    rulecolor=\color{gray!50},
    framesep=3pt,
    frameround=tttt,
    numbersep=6pt,
    xleftmargin=10pt,
    xrightmargin=10pt,
    aboveskip=1.0\baselineskip,
    belowskip=1.0\baselineskip,
    upquote=true,
    columns=flexible,
    keepspaces=true,
    mathescape=true,
    escapeinside={(*@}{@*)},
    morecomment=[l]\#,
    morekeywords={import, from, as, def, class, return, yield, for, while, if, elif, else, try, except, finally, with, lambda, pass, break, continue, and, or, not, is, in, raise, assert},
    emph={self, None, True, False, np, pd, plt, torch, tf, sklearn},
    emphstyle=\color{orange!80!black}\bfseries,
    literate=
        {-}{-}1
        {=>}{$\Rightarrow$ }3
        {->}{$\rightarrow$ }3
        {...}{$\ldots$ }3,
    inputencoding=utf8,
    extendedchars=true,
    lineskip=-0.1pt,
    fontadjust=true
}

\title{\our{}: Trajectory Prediction Heuristics Design via LLM-driven Evolution}

\author {
    Zhikai Zhao\textsuperscript{\rm 1}\equalcontrib,
    Chuanbo Hua\textsuperscript{\rm 1, 2}\equalcontrib,
    Federico Berto\textsuperscript{\rm 1, 2}\equalcontrib,
    Kanghoon Lee\textsuperscript{\rm 1},
    Zihan Ma\textsuperscript{\rm 1},\\
    Jiachen Li\textsuperscript{\rm 3},
    Jinkyoo Park\textsuperscript{\rm 1, 2}
}
\affiliations {
    \textsuperscript{\rm 1}KAIST, Korea Advanced Institute of Science and Technology, Daejeon, 34141, South Korea \\
    \textsuperscript{\rm 2}OMELET, Daejeon, South Korea\\
    \textsuperscript{\rm 3}University of California, Riverside, CA, USA\\
    \{zzk020202, cbhua, fberto, leehoon, zihanma\}@kaist.ac.kr; 
    jiachen.li@ucr.edu; 
    jinkyoo.park@kaist.ac.kr
}

\begin{document}

\maketitle

\begin{abstract}
Trajectory prediction is a critical task in modeling human behavior, especially in safety-critical domains such as social robotics and autonomous vehicle navigation. Traditional heuristics based on handcrafted rules often lack accuracy and generalizability. 
Although deep learning approaches offer improved performance, they typically suffer from high computational cost, limited explainability, and, importantly, poor generalization to out-of-distribution (OOD) scenarios. 
In this paper, we introduce \our{}, a framework that leverages Large Language Models (LLMs) to automatically design trajectory prediction heuristics. \our{} employs an evolutionary algorithm to generate and refine prediction heuristics from past trajectory data. We propose two key innovations: Cross-Generation Elite Sampling to encourage population diversity, and a Statistics Feedback Loop that enables the LLM to analyze and improve alternative predictions. 
Our evaluations demonstrate that \our{} outperforms existing heuristic methods across multiple real-world datasets, and notably surpasses both heuristic and deep learning methods in generalizing to an unseen OOD real-world dataset. \our{} marks a promising step toward the automated design of fast, explainable, and generalizable trajectory prediction heuristics. We release our source code to facilitate future research at \url{https://github.com/ai4co/trajevo}.
\end{abstract}

\section{Introduction}

Trajectory prediction is a cornerstone of intelligent autonomous systems \citep{madjid2025trajectory,wang2025deployable,wang2025uniocc}, with numerous real-world applications, including autonomous driving \citep{wang2024escirl,li2024adaptive,wang2025generative,lange2024self,wang2025cmp,lange2024scene}, industrial safety \citep{8107677}, robotic navigation \citep{vishwakarma2024adoption,li2024multi,yao2024sonic}, and planning \citep{li2022human,li2023game}. The inherent stochasticity of these environments demands systems that can accurately process data with both temporal and spatial precision \citep{li2021spatio,nakamura2024not,toyungyernsub2024predicting}. For instance, in autonomous vehicle navigation, accurately predicting pedestrian trajectories is essential for avoiding collisions in complex urban settings \citep{cao2021spectral,li2023pedestrian,choi2021shared}. Similarly, in industrial and indoor environments, robots must generate precise trajectory predictions to collaborate safely with humans in shared workspaces without causing harm \citep{mavrogiannis2023core,mahdi2022survey,ma2022multi}.

\begin{figure}[t]
    \centering
\includegraphics[width=\columnwidth]{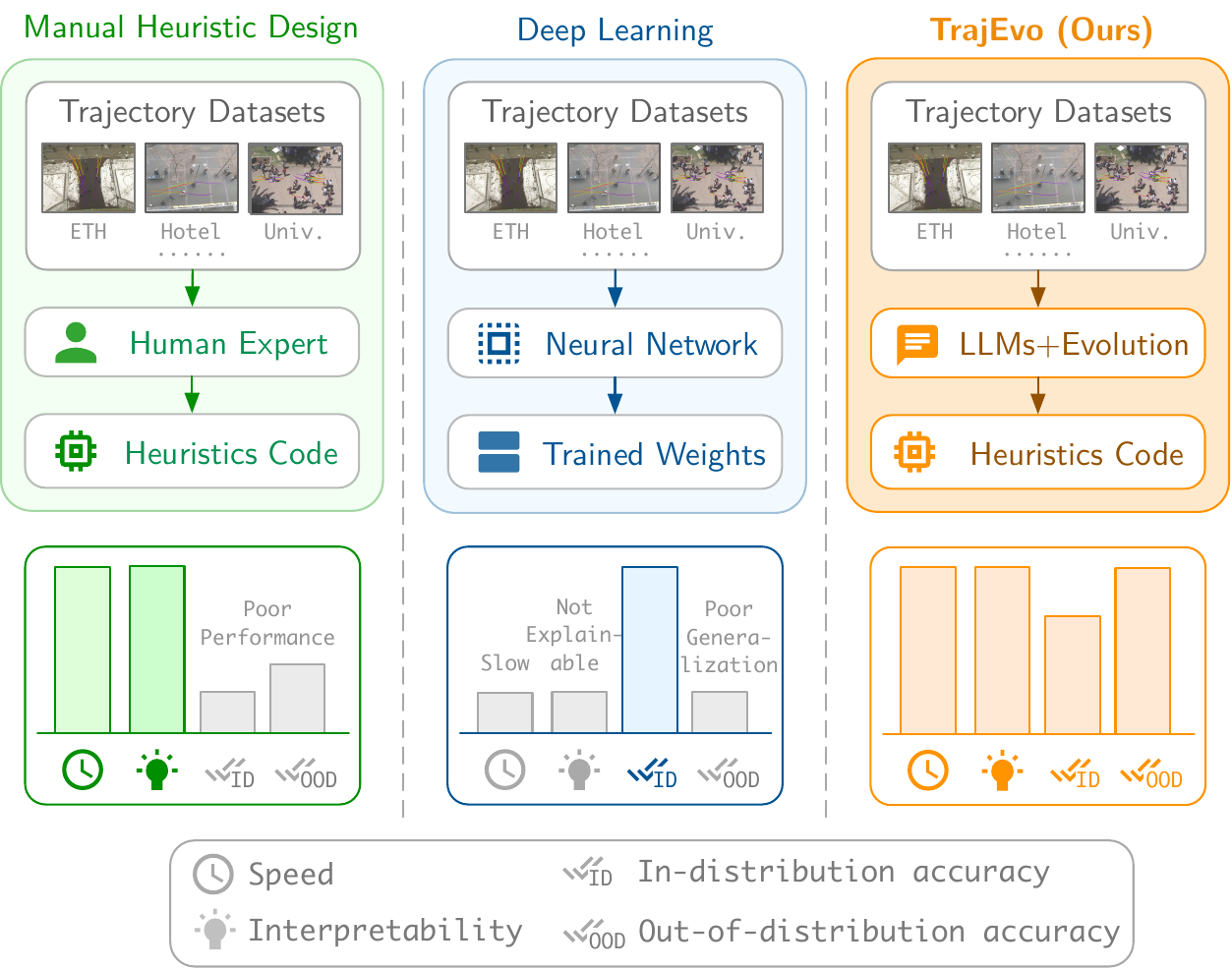}
    \caption{Motivation for our \our{} framework. Traditional manual heuristic design (left) relies on human expertise and trial-and-error processes. Deep learning approaches (center) produce more accurate predictions but demand substantial computational resources, yield black-box models, and often struggle with generalization. \our{} (right) automates the heuristic design process using evolutionary algorithms, enabling the generation of novel and effective trajectory prediction heuristics.}
    \label{fig:motivation}
\end{figure}

Accurately predicting the movement patterns of multiple agents, such as pedestrians or vehicles, remains fundamentally challenging. Human motion is inherently complex, characterized by nonlinear behaviors, sudden directional changes, and spontaneous decisions influenced by individual factors such as habits or urgency \citep{li2020evolvegraph,xu2024matrix}. Moreover, agents interact dynamically, adjusting their paths to avoid collisions, responding to traffic, or moving cohesively in groups, which results in high stochasticity \citep{amirian2020opentraj,zhou2022grouptron,li2022evolvehypergraph}.
Early efforts to model these behaviors relied on heuristic methods, including the Social Force model \citep{helbing1995social}, which captures interactions through physics-inspired forces \citep{helbing1995social,5509779,zanlungo2011social,farina2017walking,chen2018social}; constraint-based approaches that define collision-free velocities \citep{4543489,Ma2018EfficientRC}; and agent-based simulations of decision making processes \citep{5995468}.
While these handcrafted heuristics offer interpretability, they are often difficult to tune for dynamic scenarios and tend to suffer from limited accuracy and generalization.

Deep learning methods have emerged as a powerful alternative to traditional heuristics \citep{alahi2016social,salzmann2020trajectron++,li2022evolvehypergraph}. Social-LSTM \citep{alahi2016social} was among the earliest influential models, leveraging LSTMs for trajectory prediction. Since then, a wide range of approaches, such as graph neural networks (GNNs) \citep{rainbow2021semantics,ma2021multi} and generative adversarial networks (GANs) \citep{gupta2018social,li2019conditional,dendorfer2021mg}, have been proposed to further improve prediction accuracy. More recently, Transformer-based frameworks \citep{kim2025guide,bae2024can} and diffusion models \citep{yang2024uncovering,gu2022stochastic,fu2025moflow} have shown promise in capturing complex dependencies in human motion patterns.
However, deep learning methods suffer from several practical limitations. \textit{a) Running Speed:} They often require significant computational resources and exhibit high latency, making them unsuitable for deployment on resource-constrained robots or vehicles \citep{itkina2023interpretable,jiang2025survey}. 
\textit{b) Interpretability:} As black-box models, they lack transparency, which hinders verification and reduces trust, especially in safety-critical applications \citep{dax2023disentangled,liu2024traj,cai2024pwto}. \textit{c) Generalization Ability:} Perhaps most critically, they often generalize poorly in out-of-distribution (OOD) scenarios, potentially leading to unsafe behavior in unfamiliar environments \citep{korbmacher2022review,rudenko2020human}. 
These challenges highlight the continued importance of robust and predictable heuristics in safety-critical systems \citep{phong2023truly}, which requires methods that are not only accurate but also fast, explainable, and generalizable.

Therefore, we pose the following research question: \emph{Can we automatically design computationally efficient, accurate, interpretable, and highly generalizable trajectory prediction heuristics?}
Inspired by recent research exploring combinations of Large Language Models (LLMs) with Evolutionary Algorithms (EAs) for automated algorithm design \citep{dat2025hsevo,ye2024reevo,chen2024uber,liu2024evolution,yuksekgonul2025optimizing,liu2024llm4ad,wu2024evolutionary,zhang2024understanding, zheng2025monte,novikov2025alphaevolve}, we propose a novel approach to automatically generate effective prediction heuristics. We hypothesize that coupling the generative and reasoning capabilities of LLMs with the structured search of EAs can overcome the limitations of manual heuristic design to discover novel, high-performance heuristics suitable for real-world deployment.

In this paper, we introduce \our{} (\textbf{Traj}ectory \textbf{Evo}lution), a novel framework designed to achieve this goal. \our{} leverages LLMs within an evolutionary loop to iteratively generate, evaluate, and refine prediction heuristics directly from data. Our main contributions are as follows:
\begin{itemize}
    \item We present \our{}, the first framework, to the best of our knowledge, that integrates LLMs with evolutionary algorithms specifically for the automated discovery and design of fast, explainable, and robust trajectory prediction heuristics for real-world applications.
    \item We introduce a Cross-Generation Elite Sampling strategy to maintain population diversity and a Statistics Feedback Loop that enables the LLM to analyze heuristic performance and guide the generation of improved candidates based on past trajectory data.
    \item We demonstrate that \our{} generates heuristics that significantly outperform prior heuristic methods on public-open real-world datasets and exhibit remarkable generalization, achieving an over 20\% performance improvement on an unseen OOD dataset against both traditional heuristics and deep learning methods, while remaining computationally fast and interpretable.

\end{itemize}
\begin{figure*}[t] 
    \centering
    \includegraphics[width=0.8\textwidth]{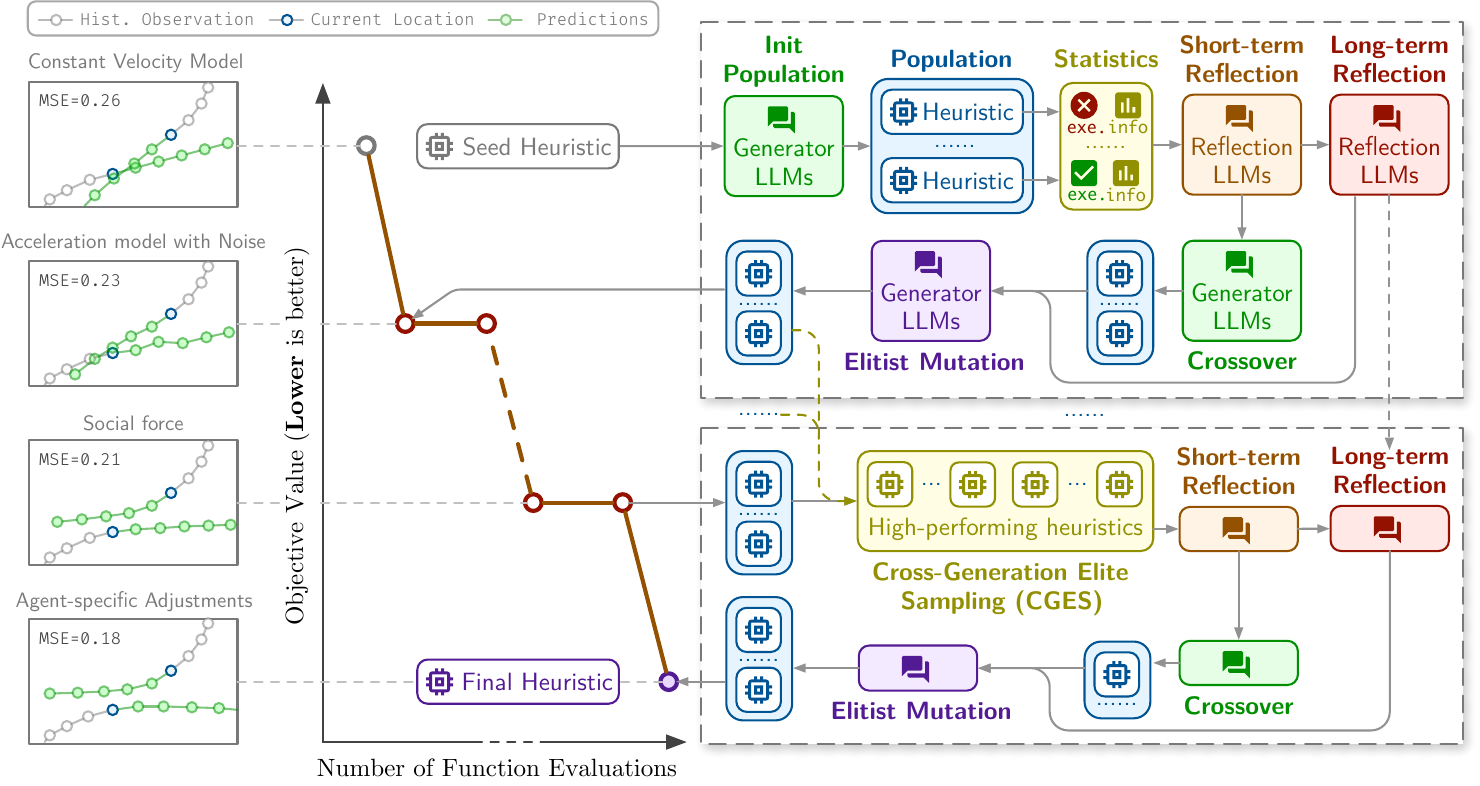}
    \vspace{-0.2cm}
    \caption{An illustration of the \our{} evolutionary process. \textbf{(Left)} The framework continuously discovers and evaluates a variety of heuristic strategies, such as those based on constant velocity or social forces. \textbf{(Middle)} As the evolution progresses, the performance of these heuristics steadily improves, evidenced by the decreasing objective value from a simple seed to a final, optimized solution. \textbf{(Right)} The overall \our{} pipeline orchestrates this entire discovery and optimization process, using an LLM-driven evolutionary algorithm to automatically generate and refine the heuristics.}
    \vspace{-0.2cm}
    \label{fig:trajevo-catchy}
\end{figure*}

\section{Related Work}

\paragraph{Heuristic Methods for Trajectory Prediction}
Traditional heuristic approaches provide interpretable frameworks for trajectory prediction, but often with accuracy limitations. The Constant Velocity Model (CVM) and its sampling variant (CVM-S) \citep{scholler2020constant} represent foundational baselines that assume uniform motion patterns. More sophisticated approaches include Constant Acceleration \citep{polychronopoulos2007sensor}, CTRV (Constant Turn Rate and Velocity) \citep{lu2021ctrv}, and CSCRCTR \citep{s140305239}, which incorporate different kinematic assumptions. The Social Force Model \citep{helbing1995social} pioneered physics-inspired representations of pedestrian dynamics, with numerous extensions incorporating social behaviors \citep{zanlungo2011social, farina2017walking, chen2018social} and group dynamics \citep{helbing1997modeling}. Despite their efficiency and explainability, they typically struggle with complex real-world, multi-agent interactions due to their limited expressiveness and reliance on manually defined parameters -- limitations our proposed \our{} framework specifically addresses through automatic heuristic generation.

\paragraph{Learning-based Trajectory Prediction}
Deep learning has substantially advanced trajectory prediction accuracy at the cost of computational efficiency and interpretability \citep{sun2022interaction,xie2023cognition}. Social-LSTM \citep{alahi2016social} is a seminal work using a social pooling mechanism to model inter-agent interactions, while Social-GAN \citep{gupta2018social} leveraged generative approaches to capture trajectory multimodality. Graph-based architectures like STGAT \citep{Huang_2019_ICCV} and Social-STGCNN \citep{mohamed2020social} explicitly model the dynamic social graph between pedestrians. Recent approaches include Trajectron++ \citep{salzmann2020trajectron++}, which integrates various contextual factors, MemoNet \citep{xu2022remember} with its memory mechanisms, EigenTrajectory \citep{bae2023eigentrajectory} focusing on principal motion patterns, Hyper-STTN \citep{wang2024hyper} utilizing hypergraph-based spatial-temporal transformers for group-aware trajectory prediction, and MoFlow \citep{fu2025moflow} employing normalizing flows for stochastic prediction. 
While these methods achieve high in-distribution accuracy, their three well-documented challenges: high computational cost, lack of interpretability, and poor out-of-distribution generalization \citep{rudenko2020human}, motivate the design of our \our{} framework.

\paragraph{LLMs for Algorithmic Design}
Recent work has combined Large Language Models (LLMs) with Evolutionary Algorithms (EAs) for automated algorithm design \citep{ye2024reevo, dat2025hsevo}. However, applying this paradigm to the safety-critical and stochastic domain of trajectory prediction remains unexplored. 
To the best of our knowledge, \our{} is the first framework to adapt LLM-driven EAs for discovering fast, interpretable, and generalizable trajectory prediction heuristics.
While these studies establish the general potential of LLM-driven evolution, its application to specialized, safety-critical domains remains largely unexplored. The domain of multi-agent trajectory prediction presents unique challenges, including stochastic interactions and the need for both high accuracy and computational efficiency. To the best of our knowledge, \our{} is the first framework to adapt the LLM-driven evolutionary paradigm specifically for this task. 

\section{\our{}}

\subsection{Problem Definition}
\label{subsec:problem-definition}

\paragraph{Task}
We address multi-agent trajectory prediction. The task is, for each agent $i$, to predict its future path $Y_i = (P_i^{T_{\text{obs}}+1}, \dots, P_i^{T_{\text{obs}}+T_{\text{pred}}})$ given its observed history $H_i = (P_i^1, \dots, P_i^{T_{\text{obs}}})$, where $P_i^t \in \mathbb{R}^2$ is the position at timestep $t$. Here, $T_{\text{obs}}$ is the historical observation span and $T_{\text{pred}}$ is the future prediction length.

\paragraph{Metrics}
We adopt the widely used standard measurement matrix in trajectory prediction to ensure fair comparison with all baselines \citep{gupta2018social}. Performance is measured by the minimum Average Displacement Error ($\text{minADE}_{K}$) and minimum Final Displacement Error ($\text{minFDE}_{K}$). 
This matrix evaluates a model's ability to capture the multi-modal nature of human motion by generating $K$ trajectory samples and selecting the one with the lowest error against the ground truth. Following the same setting as the baselines, we use $K=20$ for our method to ensure a fair and direct comparison.

\paragraph{Optimization Objective}
To guide the evolutionary search, we employ the Mean Squared Error (MSE) and calculate the average squared Euclidean distance between the predicted and ground truth trajectory points, where a lower MSE corresponds to a higher fitness. 
The choice of MSE is motivated by its simplicity and generality, as well as its capacity to provide a continuous and well-defined error signal, which is crucial for guiding the evolutionary process.
Its widespread adoption ensures our results are comparable across the literature, while its parameter-free nature aligns with our goal of automated heuristic design.

\subsection{Evolutionary Framework}
\label{subsec:evolutionary-framework} 

Our evolutionary framework is designed not merely to minimize prediction error on a given dataset, but to discover heuristics that are fundamentally simple and robust enough to generalize across different environments. Inspired by the Reflective Evolution approach \citep{ye2024reevo}, the framework is illustrated in \cref{fig:trajevo-catchy} (right).
In this algorithm, Large Language Models (LLMs) serve as the core genetic operators, following these steps:
\paragraph{Initial Population}
The process starts by seeding the generator LLM with a task specification (including problem details, input/output format, and the objective $J$) alongside a basic heuristic like the Constant Velocity Model (CVM) \citep{scholler2020constant}. The LLM then generates an initial population of $N$ diverse heuristics, providing the starting point for the evolutionary search.

\paragraph{Selection for Crossover}
Parents for crossover are chosen from successfully executed heuristics in the current population. The selection balances exploration and exploitation: 70\% of parents are selected uniformly at random from successfully running candidates, while 30\% are chosen from the elite performers (those with the lowest objective $J$).

\paragraph{Reflections}
\our{} employs two types of reflection as in \citet{ye2024reevo}. \textit{Short-term reflections} compare the performance of selected crossover parents, offering immediate feedback to guide the generation of offspring. \textit{Long-term reflections} accumulate insights across generations, identifying effective design patterns and principles to steer mutation and broader exploration. Both reflection mechanisms produce textual guidance (i.e. ``verbal gradients'' \citep{pryzant2023automatic}) for the generator LLM.

\paragraph{Crossover}
This operator creates new offspring by combining code from two parent heuristics. Guided by short-term reflections that compare the `better' and `worse' performing parent, the LLM is prompted to mix their effective ``genes", enabling the emergence of potentially superior heuristics.

\paragraph{Elitist Mutation} 
The mutation operator mutates the elitist (best found so far) heuristic. In \our{}, this involves the generator LLM modifying an elite heuristic selected. This mutation step is informed by the insights gathered through long-term reflections.

It is crucial to note that this entire evolutionary process is a one-time, offline procedure. The final output is a standalone Python heuristic that runs with high computational efficiency and does not require the LLM during inference.

\subsection{Cross-Generation Elite Sampling}
\label{subsec:cross-generation-elite-sampling}
\begin{figure}[h!]
    \centering
    \includegraphics[width=0.92\columnwidth]{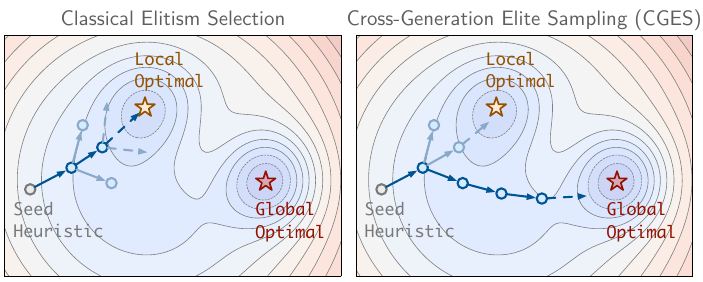}
    \caption{Cross-Generation Elite Sampling (CGES) helps escape local optima by sampling elite individuals from past generations (left), which greatly helps achieve much better objective values (right).}
    \label{fig:cross-generation-elite-sampling}
\end{figure}
Evolving effective heuristics for complex tasks like trajectory prediction poses a significant search challenge, where standard evolutionary processes can easily get trapped in local optima \citep{osuna2018runtime}. Simple mutations often yield only incremental improvements, failing to explore sufficiently diverse or novel strategies. To address this limitation and enhance exploration, we introduce Cross-Generation Elite Sampling (CGES), a core component influencing the mutation step within the \our{} framework.

In contrast to typical elitism focusing on the current generation's best, CGES maintains a history archive of high-performing heuristics accumulated across all past generations. Specifically, CGES modifies how the elite individual targeted for mutation is selected: instead of necessarily choosing the top performer from the current population, the heuristic designated to undergo mutation is sampled by CGES directly from this history. This sampling uses a Softmax distribution based on the recorded objectives $J$ of the historical elites, thus prioritizing individuals that have proven effective. By potentially reintroducing and modifying diverse, historically successful strategies during the mutation phase, CGES significantly improves the exploration capability, facilitating escape from local optima and the discovery of more robust heuristics, as demonstrated in \cref{fig:cross-generation-elite-sampling}.

\subsection{Statistics Feedback Loop}
\label{subsec:statistics-feedback-loop}
\begin{figure}[!tbp]
    \centering 
    \begin{subfigure}{0.8\columnwidth}
        \centering
        \includegraphics[width=\textwidth]{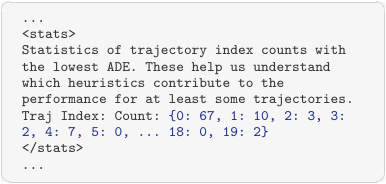} 
        \label{fig:stats_analysis}
    \end{subfigure}
    \begin{subfigure}{0.84\columnwidth}
        \centering
        \includegraphics[width=\textwidth]{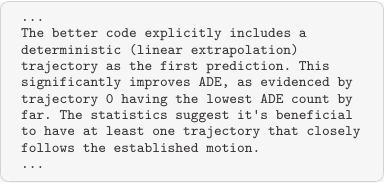} 
        \label{fig:stats_refinement}
    \end{subfigure}
    \caption{Statistics Feedback Loop: (top) distribution of prediction index effectiveness by minimum ADE frequency; (bottom) LLM prompt incorporating statistical feedback for guided mutation.}
    \label{fig:stats_feedback_loop} 
\end{figure}

While the objective $J$ measures overall quality, it does not reveal \textit{which} of a heuristic's diverse internal prediction strategies are actually effective. To provide this crucial insight for effective refinement, \our{} incorporates a Statistics Feedback Loop (SFL), illustrated in \cref{fig:stats_feedback_loop}.

This loop specifically analyzes the contribution of the 20 distinct trajectory prediction sets (indexed $k=0\dots19$) generated by a heuristic. After evaluation, we compute a key statistic: a distribution showing how frequently each prediction index $k$ delivered the minimum ADE for individual trajectory instances across the dataset (\cref{fig:stats_feedback_loop} top). This directly highlights the empirical utility of the different diversification strategies associated with each index $k$. This statistical distribution, together with the heuristic's code, is fed to the reflector and mutation LLMs to identify which strategies ($k$) contribute most effectively to performance (\cref{fig:stats_feedback_loop} bottom). Such feedback provides actionable guidance to the generator LLM, enabling specific improvements to the heuristic's multi-prediction generation logic based on the observed effectiveness of its constituent strategies. A qualitative example of the full \our{} evolution is provided in \cref{fig:trajevo-catchy}.

\begin{table*}[t]
\centering
\small
\setlength{\tabcolsep}{1mm} 
\begin{tabular}{l||c|c|c|c|c||c}
\toprule
Method & ETH & HOTEL & UNIV & ZARA1 & ZARA2 & AVG \\
\midrule
SocialForce \cite{helbing1995social} & 1.46/2.48 & 0.69/1.23 & 0.96/1.75 & 1.37/2.51 & 0.84/1.53 & 1.06/1.90 \\
LinReg \citep{bishop2006pattern} &  1.04/2.20 & 0.26/0.47 & 0.76/1.48 & 0.62/1.22 & 0.47/0.93 & 0.63/1.26 \\
ConstantAcc \citep{polychronopoulos2007sensor}  & 3.12/7.98 & 1.64/4.19 & 1.02/2.60 & 0.81/2.05 & 0.60/1.53 & 1.44/3.67 \\
CSCRCTR \cite{s140305239} & 2.27/4.61 & 1.03/2.18 & 1.35/3.12& 0.96/2.12 & 0.90/2.10  & 1.30/2.83 \\
CVM \cite{scholler2020constant}   & 1.01/2.24 & 0.32/0.61 & 0.54/1.21 & 0.42/0.95 & 0.33/0.75 & 0.52/1.15 \\
CVM-S \cite{scholler2020constant}  & 0.92/2.01 & 0.27/0.51 & 0.53/1.17 & 0.37/0.77 & 0.28/0.63 & 0.47/1.02 \\
CTRV \cite{lu2021ctrv} & 1.62/3.64 & 0.72/1.09 & 0.71/1.59 & 0.65/1.50 & 0.48/1.10 & 0.84/1.78 \\

\midrule
\our{}  
& \textbf{0.47/0.77} 
& \textbf{0.17/0.30} 
& \textbf{0.51/1.10} 
& \textbf{0.35/0.75} 
& \textbf{0.27/0.57} 
& \textbf{0.35/0.70} \\
\bottomrule
\end{tabular}
\caption{Comparison of \our{} with trajectory prediction heuristics across datasets with mean minADE$_{20}$ / minFDE$_{20}$ (meters) on the ETH-UCY dataset.}
\label{table:main_hs}
\end{table*}

\begin{table*}[t]
\centering
\small
\setlength{\tabcolsep}{1mm} 
\begin{tabular}[\linewidth]{l||c|c|c|c|c||c}
\toprule
Method & ETH & HOTEL & UNIV & ZARA1 & ZARA2 & AVG \\
\midrule
Social-LSTM \cite{alahi2016social}
 & \underline{1.09/2.35}
 & \underline{0.79/1.76}
 & \underline{0.67/1.40}
 & \underline{0.56/1.17}
 & \underline{0.72/1.54}
 & \underline{0.77/1.64} \\

Social-GAN \cite{gupta2018social}
 & \underline{0.87/1.62}
 & \underline{0.67/1.37}
 & \underline{0.76/1.52}
 & 0.35/0.68
 & \underline{0.42/0.84}
 & \underline{0.61/1.21} \\

STGAT \cite{Huang_2019_ICCV}
 & \underline{0.65/1.12}
 & \underline{0.35/0.66}
 & \underline{0.52/1.10}
 & 0.34/0.69
 & \underline{0.29/0.60}
 & \underline{0.43/0.83} \\

Social-STGCNN \cite{mohamed2020social}
 & \underline{0.64/1.11}
 & \underline{0.49/0.85}
 & 0.44/0.79
 & 0.34/0.53
 & \underline{0.30/0.48}
 & \underline{0.44/0.75} \\

Trajectron++ \cite{salzmann2020trajectron++}
 & \underline{0.61/1.03}
 & \underline{0.20}/0.28
 & 0.30/0.55
 & 0.24/0.41
 & 0.18/0.32
 & 0.31/0.52 \\

MemoNet \cite{xu2022remember}
 & 0.41/0.61
 & \textbf{0.11}/\textbf{0.17}
 & 0.24/0.43
 & 0.18/0.32
 & 0.14/0.24
 & 0.21/0.35 \\

EigenTrajectory \cite{bae2023eigentrajectory}
 & \textbf{0.36}/\textbf{0.53}
 & 0.12/0.19
 & 0.24/0.43
 & 0.19/0.33
 & 0.14/0.24
 & 0.21/0.34 \\

MoFlow \cite{fu2025moflow}
 & 0.40/0.57
 & \textbf{0.11}/\textbf{0.17}
 & \textbf{0.23}/\textbf{0.39}
 & \textbf{0.15}/\textbf{0.26}
 & \textbf{0.12}/\textbf{0.22}
 & \textbf{0.20}/\textbf{0.32} \\

\midrule
\our{}
& 0.47/0.77
& 0.17/0.30 
& 0.51/1.10 
& 0.35/0.75
& 0.27/0.57
& 0.35/0.70 \\
\bottomrule
\end{tabular}
\caption{Comparison of \our{} with deep learning approaches (mean minADE$_{20}$/minFDE$_{20}$ on ETH-UCY). Each \underline{underlined} number indicates that the result is worse than the corresponding result from \our{} on at least one metric.}
\label{table:main_nn}
\end{table*}

\section{Experiments}
\label{sec:result}

\subsection{Experimental Setup}
\paragraph{Datasets}
We evaluate \our{} on the ETH-UCY benchmark \citep{Pellegrini_2009_ICCV,lerner2007crowds}, a collection of datasets comprising real-world pedestrian trajectories with the standard leave-one-out protocol where heuristics are evolved on four datasets and tested on the remaining one \citep{alahi2016social}. 
To specifically assess OOD generalization, we use the SDD dataset \citep{robiczek2016learning} as a completely unseen test set.
For all experiments, following the standard setting widely used by baseline methods on these benchmarks, we observe 8 past frames (3.2s) to predict 12 future frames (4.8s).

\paragraph{Baselines}
We compare heuristics generated by \our{} against both heuristic and deep learning baselines.
\textit{Heuristic baselines}, which are well-suited for resource-constrained systems, include kinematic models such as the Constant Velocity Model (CVM) and its sampling variant (CVM-S) \citep{scholler2020constant}, Constant Acceleration (ConstantAcc) \citep{polychronopoulos2007sensor}, and Constant Turn Rate and Velocity (CTRV) \citep{lu2021ctrv}, as well as CSCRCTR \citep{s140305239}, Linear Regression (LinReg) \citep{bishop2006pattern}, and the physics-inspired Social Force model \citep{helbing1995social}.
To benchmark against more complex data-driven approaches, we also evaluate \textit{deep learning baselines}, including early influential models like Social-LSTM \citep{alahi2016social} and Social-GAN \citep{gupta2018social}; graph-based methods such as STGAT and Social-STGCNN \citep{mohamed2020social}; and recent state-of-the-art models like Trajectron++ \citep{salzmann2020trajectron++}, MemoNet \citep{xu2022remember}, EigenTrajectory \citep{bae2023eigentrajectory}, and MoFlow \citep{fu2025moflow}.
To ensure robustness, all reported results for \our{} are averaged over 10 independent evolutionary runs. The standard deviations across runs were consistently low, typically ranging from 0.02 to 0.06.

\paragraph{Hardware and Software}
All experiments were conducted on a workstation equipped with an AMD Ryzen 9 7950X 16-Core Processor and a single NVIDIA RTX 3090 GPU. The \our{} framework generates trajectory prediction heuristics as executable Python code snippets in a Python 3.12 environment, employing Google's Gemini 2.0 Flash model \citep{GoogleAIDev_GeminiPricing_2025} in the main experiment. To evaluate the stability and generalizability of our framework, its performance was systematically tested with a variety of alternative models, including DeepSeek V3 \& R1, Qwen3-32B, ChatGPT-4o, and Claude 3.7 Sonnet.

\subsection{In-Distribution Performance}

\begin{figure}[!tbp]  
    \centering
    \includegraphics[width=\columnwidth]{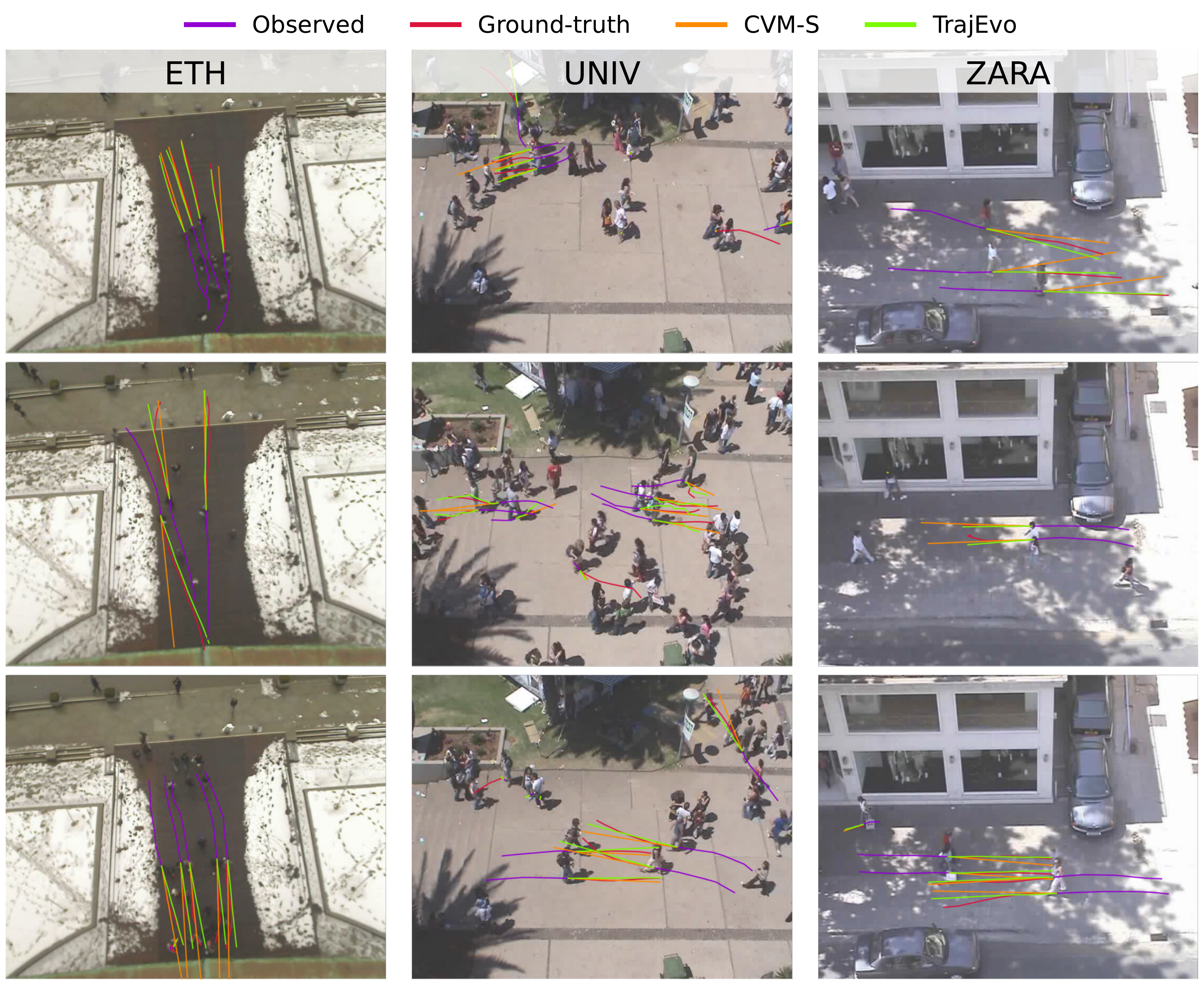}
    \caption{Trajectory prediction results for CVM-S \citep{scholler2020constant} and \our{} across different datasets. Each row illustrates a distinct pattern: (top) linear trajectories, (middle) non-linear trajectories, and (bottom) collision-avoidance cases. 
    We visualize the single best trajectory out of $K=20$ samples based on the optimization objective.}
    \vspace{-0.3cm}
    \label{fig:generated-trajectories-image}
\end{figure}

\paragraph{Comparison with Heuristic Methods}
We report results against heuristic baselines in \cref{table:main_hs}. \our{} consistently achieves the best performance across all ETH-UCY datasets and significantly outperforms all competitors on average. This establishes \our{} as the new state-of-the-art among heuristic approaches on this benchmark.  Interestingly, the general performance trend across baseline methods suggests that heuristics incorporating more complexity beyond basic constant velocity assumptions are generally worse suited for real-world pedestrian data, including SocialForce, with the second-best result obtained by the relatively simple CVM-S. \cref{fig:generated-trajectories-image} shows examples of the generated trajectories. 

\begin{table*}[t]
\centering
\small
\setlength{\tabcolsep}{1mm} 
\begin{tabular}{l c c c c c c}
\toprule
Method (tested on SDD dataset) & ETH & HOTEL & UNIV & ZARA1 & ZARA2 & AVG \\
\midrule
SocialForce \cite{helbing1995social} 
& 33.64/60.63 & 33.64/60.63 & 33.64/60.63 & 33.64/60.63 & 33.64/60.63 & 33.64/60.63 \\
CVM \cite{scholler2020constant}
  & 18.82/37.95 & 18.82/37.95 & 18.82/37.95 & 18.82/37.95 & 18.82/37.95 & 18.82/37.95\\
CVM-S \cite{scholler2020constant}
  & 16.28/31.84 & 16.28/31.84 & 16.28/31.84 & 16.28/31.84 & 16.28/31.84 & 16.28/31.84 \\
  \midrule
Trajectron++ \cite{salzmann2020trajectron++}
  & 46.72/69.11 & 47.30/67.76 & 46.08/75.90 & 47.30/72.19 & 46.78/68.59 & 46.84/70.71 \\
EigenTrajectory \cite{bae2023eigentrajectory}
  & 14.51/25.13 & 14.69/24.64 & 14.31/27.60 & 14.69/26.25 & 14.53/24.94 & 14.55/25.71 \\
 MoFlow \cite{fu2025moflow}
& 17.00/27.98 & 17.21/27.43 & 17.00/30.63 & 17.24/29.22 & 17.27/27.56 & 17.14/28.56 \\
  \midrule
\our{}
  & \textbf{12.58}/\textbf{23.82} & \textbf{12.26}/\textbf{23.60} & \textbf{12.78}/\textbf{23.71} & \textbf{13.10}/\textbf{25.23} & \textbf{12.22}/\textbf{23.57} & \textbf{12.59}/\textbf{23.99} \\
\bottomrule
\end{tabular}
\caption{Out of distribution performance of methods trained on different ETH-UCY splits and tested on the unseen SDD dataset. We report minADE$_{20}$ / minFDE$_{20}$ (pixels) on the SDD dataset.}
\label{tab:cross-dataset-generalization}
\end{table*}

\paragraph{Comparison with Deep Learning Methods} 
\cref{table:main_nn} reports results against deep learning methods. On this in-distribution benchmark, while \our{} does not surpass highly specialized neural networks like MoFlow \citep{fu2025moflow}, it establishes a strong baseline. It consistently outperforms several established models, including Social-LSTM \citep{alahi2016social}, and even more recent ones like Trajectron++ \citep{salzmann2020trajectron++} on certain datasets. This performance is highly competitive for a heuristic-based approach. However, as the next section demonstrates, the primary strength of our method lies not in incremental in-distribution gains, but in its superior generalization capabilities in OOD situations.

\subsection{Out-of-Distribution Performance}
A crucial capability for robotic systems deployed in the real world is generalizing to unseen (OOD) environments during development. 
We evaluate this on the unseen dataset SDD, directly testing both deep learning models (trained on ETH-UCY) and the heuristics from \our{} (evolved on the ETH-UCY).
We report these OOD performance results in \cref{tab:cross-dataset-generalization} for three selected heuristics and three recent established neural methods. Remarkably, \our{} demonstrates superior generalization, significantly outperforming not only all heuristic baselines but also all tested deep learning methods, including SOTA models like MoFlow \citep{fu2025moflow}. \our{} performs substantially better than the best deep learning competitor, EigenTrajectory \citep{bae2023eigentrajectory}  and MoFlow. This suggests that the interpretable and efficient heuristics discovered by \our{} may possess greater robustness to domain shifts compared to complex neural networks trained on specific distributions. For a bidirectional validation of generalization ability, we provide the reverse experiment in the Appendix.
\subsection{Ablation Study} 
We conducted an ablation study, reported in \cref{tab:ablation}, to validate the effectiveness of the core components introduced in \our{}. The results demonstrate that both the SFL and CGES meaningfully improve performance. Removing either component from the full framework leads to a noticeable degradation in prediction accuracy, confirming their positive contribution. The significant impact of CGES on enhancing the quality of generated heuristics during evolution is further visualized in Appendix \cref{fig:cges_ablation_appendix}. 

\begin{table}[h!]
\centering
\small
\begin{tabular}{@{}lccc@{}}
\toprule
\setlength{\tabcolsep}{1mm} 
{Dataset} & {~ - SFL - CGES} & {~ - SFL} & {\our{}} \\
\midrule
ETH   & 0.68/1.36 & 0.59/1.12 & \textbf{0.47/0.77} \\
HOTEL & 0.26/0.45 & 0.19/0.33 & \textbf{0.17/0.30} \\
UNIV  & 0.59/1.22 & 0.52/1.13 & \textbf{0.51/1.10} \\
ZARA1 & 0.37/0.77 & 0.36/0.76 & \textbf{0.35/0.75} \\
ZARA2 & 0.31/0.65 & 0.28/0.59 & \textbf{0.27/0.57} \\
\bottomrule
\end{tabular}
\caption{Ablation study for the evolution framework removing different components. Lower is better $(\downarrow)$.}
\label{tab:ablation}
\end{table}

\subsection{Analysis and Discussion}
\paragraph{Performance of Different LLMs}
To demonstrate the robustness of the \our{} framework and the stability of its outcomes, we evaluated its performance when driven by several state-of-the-art LLMs, including DeepSeek-V3 \& R1, Qwen3-32B, ChatGPT-4o, and Claude 3.7 Sonnet. As detailed in \cref{tab:llm}, a key finding is the consistent effectiveness across all models. Each LLM was capable of generating high-performing heuristics, which underscores the general utility and stability of \our{}. While the overall performance level remained high and stable, we did observe minor variations that highlight opportunities for fine-tuning. For instance, Deepseek-R1 achieved the best average results, while other models excelled on specific datasets. This analysis confirms that the \our{} framework is effective and not reliant on a single specific LLM, ensuring stable, high-quality outcomes regardless of the chosen model.

\begin{table}[t]
\centering
\small
\setlength{\tabcolsep}{1mm} 
\begin{tabular}{l cc cc cc}
\toprule
\textbf{Model} & \multicolumn{2}{c}{\textbf{ETH}} & \multicolumn{2}{c}{\textbf{HOTEL}} & \multicolumn{2}{c}{\textbf{UNIV}} \\
\cmidrule(lr){2-3} \cmidrule(lr){4-5} \cmidrule(lr){6-7}
& ADE & FDE & ADE & FDE & ADE & FDE \\
\midrule
deepseek-R1  & 0.46 & 0.74 & 0.16 & \textbf{0.29} & \textbf{0.49} & \textbf{1.06} \\
deepseek-V3  & \textbf{0.43} & \textbf{0.71} & 0.21 & 0.39 & 0.50 & 1.08 \\
Qwen3-32B    & 0.49 & 0.74 & 0.18 & 0.32 & 0.51 & 1.09 \\
ChatGPT-4o   & 0.51 & 0.82 & \textbf{0.15} & 0.30 & 0.54 & 1.13 \\
Claude-3.7   & 0.49 & 0.93 & 0.18 & 0.36 & \textbf{0.49} & 1.07 \\
\midrule[1pt] 
\textbf{Model} & \multicolumn{2}{c}{\textbf{ZARA1}} & \multicolumn{2}{c}{\textbf{ZARA2}} & \multicolumn{2}{c}{\textbf{AVG}} \\
\cmidrule(lr){2-3} \cmidrule(lr){4-5} \cmidrule(lr){6-7}
& ADE & FDE & ADE & FDE & ADE & FDE \\
\midrule
deepseek-R1  & 0.33 & 0.72 & \textbf{0.26} & \textbf{0.54} & \textbf{0.34} & \textbf{0.67} \\
deepseek-V3  & 0.34 & \textbf{0.69} & 0.27 & 0.56 & 0.35 & 0.69 \\
Qwen3-32B    & 0.35 & 0.75 & 0.27 & 0.59 & 0.36 & 0.70 \\
GPT-4o       & 0.36 & 0.72 & 0.31 & \textbf{0.54} & 0.37 & 0.70 \\
Claude-3.7   & \textbf{0.32} & \textbf{0.61} & 0.27 & 0.57 & 0.35 & 0.71 \\
\bottomrule
\end{tabular}
\caption{Performance of various LLMs on the standard ETH-UCY benchmark datasets. Best results are in bold.}
\label{tab:llm}
\end{table}

\paragraph{Evolution Resources}
A single \our{} evolutionary run takes approximately 5 minutes using Google's Gemini 2.0 Flash \citep{GoogleAIDev_GeminiPricing_2025} with an average API cost of \$0.05. In contrast, training neural methods can take a full day on a contemporary GPU, incurring significantly higher estimated compute costs -- e.g., on an RTX 3090 as reported by \citet{bae2023eigentrajectory}, around \$4 based on typical rental rates, 80$\times$ more than \our{}.

\paragraph{Inference Resources} 
Crucially, the LLM is only used during the offline evolutionary design phase and is not called at inference time. The resulting heuristics are lightweight Python functions. As such, \our{}-generated heuristics demonstrate significant speed advantages, requiring only 0.65ms per instance on a single CPU core with the Python version and less than 5$\mu$s for a version of the heuristic rewritten in C++. In contrast, neural baselines need 12-29ms on GPU and 248-375ms on (multi-core) CPU. Notably, \our{} achieves over 10$\times$ speedup for the original Python version and more than 2400$\times$ speedup for the C++ version compared to the optimized MoFlow running on a dedicated GPU, highlighting its relevance for resource-constrained, real-time robotic systems. 

\paragraph{Explainability} 
Unlike the black-box nature of neural networks, \our{} produces human-readable Python code. For instance, the heuristic evolved for the Zara1 dataset (see Supplementary Material for the full code) is not a simple tweak of a known model. It intelligently combines four distinct strategies to generate its 20 samples. The primary strategy ($k=0..9$) is a sophisticated form of adaptive linear extrapolation, where the velocity vector is averaged over the last few steps and perturbed by noise scaled to the agent's current speed—allowing it to model both smooth paths and sudden hesitations. A second strategy ($k=10..14$) introduces rotational noise, effectively simulating curved paths. A third ($k=15..17$) implements a memory-based collision avoidance mechanism by checking potential future positions against the recent paths of nearby agents. The final strategy ($k=18,19$) acts as a conservative fallback, using a heavily damped extrapolation. This automated discovery of a complex, multi-faceted strategy combining adaptive kinematics and interaction rules is a key advantage, making the model's logic transparent and verifiable, which is crucial for safety-critical applications.

\section{Conclusion}
\label{sec:conclusion}
We introduced \our{}, a novel framework leveraging Large Language Models and evolutionary algorithms to automate the design of trajectory prediction heuristics. Our experiments demonstrate that \our{} generates heuristics which not only outperform traditional methods on standard benchmarks but also exhibit superior out-of-distribution generalization performance, remarkably surpassing even deep learning models on unseen data while remaining fast and interpretable. We believe \our{} represents a significant first step towards automatically discovering efficient, explainable, and generalizable trajectory prediction heuristics, offering a practical and powerful alternative to conventional black-box models. 

Future research will focus on bridging the performance gap to state-of-the-art models by enhancing the sophistication of the evolutionary search and incorporating multi-modal contextual inputs. To ensure greater real-world applicability, the optimization objective will be reframed from single metrics of predictive accuracy towards holistic performance on downstream tasks like navigation and planning.

\clearpage
\pagebreak 

\bibliography{aaai2026}

\clearpage

\onecolumn
\appendix

\section*{Supplementary Materials}
\section{Qualitative Example of the Evolutionary Process}
\label{appendix:evo_example}

\begin{figure*}[h!]
    \centering
    \includegraphics[width=\textwidth]{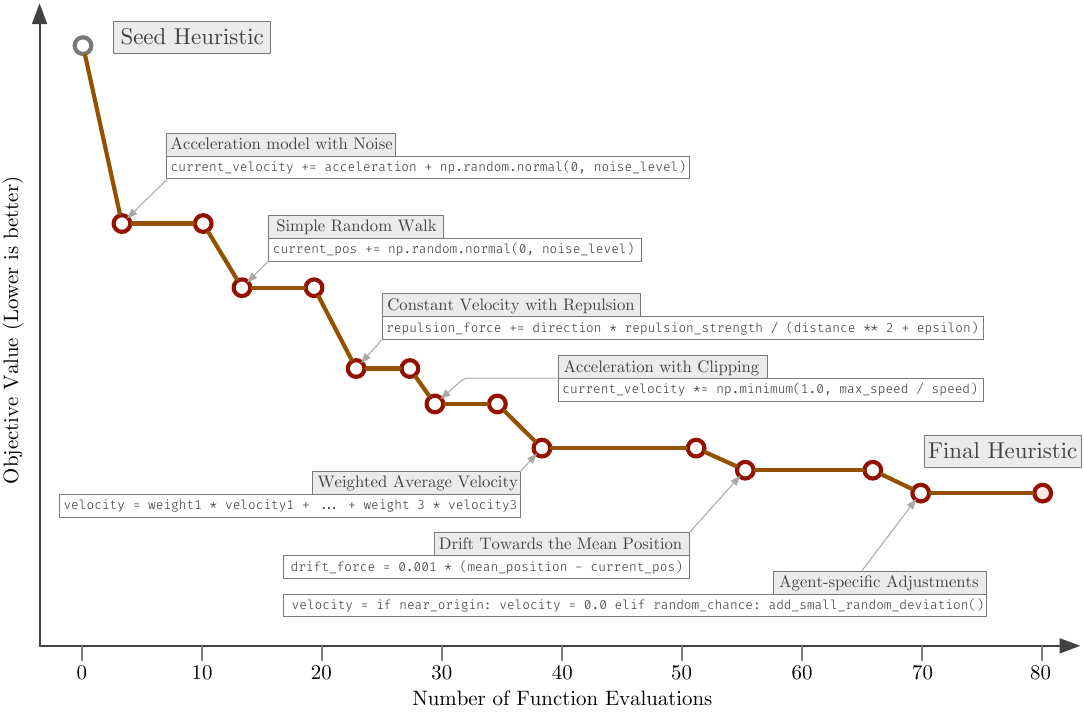}
    \caption{A qualitative illustration of the \our{} evolutionary process. The optimization begins with a simple seed heuristic and iteratively discovers more complex and effective strategies by exploring different concepts like acceleration with noise, social repulsion, and weighted velocity averaging. The process converges on a final, highly-optimized heuristic that combines multiple discovered strategies, achieving a significantly lower objective value.}
    \label{fig:appendix_evo_example}
\end{figure*}

The evolutionary process of \our{} is qualitatively illustrated in Figure~\ref{fig:appendix_evo_example}. The optimization begins with a simple \textbf{Seed Heuristic}, which has a relatively high initial objective value (lower is better). Guided by the LLM, \our{} then explores the solution space by generating and testing diverse new heuristics.

As shown in the figure, this exploration phase involves testing various distinct strategies, such as adding noise to an acceleration model (\textbf{Acceleration model with Noise}), simulating social forces (\textbf{Constant Velocity with Repulsion}), or averaging velocity over time (\textbf{Weighted Average Velocity}). Through mechanisms like the Statistics Feedback Loop and Cross-Generation Elite Sampling, the system identifies and refines the most effective of these strategies.

The process ultimately converges on a highly optimized \textbf{Final Heuristic}. This final algorithm is not a single, simple model but a complex combination of the most successful traits discovered during evolution, such as a drift towards the mean position and other agent-specific adjustments. This demonstrates the ability of \our{} to automate the design of a sophisticated, high-performance heuristic from a basic starting point.

\section{Experimental Details}

\subsection{Hyperparameters}

The \our{} framework employs several hyperparameters that govern the evolutionary search process and the interaction with Large Language Models. Key parameters used in our experiments are detailed in \cref{tab:hyperparams}.

These include settings for population management within the evolutionary algorithm, the genetic operators, LLM-driven heuristic generation and reflection mechanisms, as well as task-specific constants for trajectory prediction evaluation. Many of these settings were determined based on common practices in evolutionary computation, adaptations from the ReEvo framework \citep{ye2024reevo}, or empirically tuned for the trajectory prediction task.

\begin{table}[htbp]
    \centering
    \small 
    \caption{Main hyperparameters for the TRAJEVO framework.}
    \label{tab:hyperparams}
    \begin{tabular}{@{} p{5.5cm} l @{}}
        \toprule
        \textbf{Hyperparameter} & \textbf{Value} \\
        \midrule
        
        \multicolumn{2}{@{}l}{\textbf{Evolutionary Algorithm}} \\
        \quad Population size & 10 \\
        \quad Number of initial generation & 8 \\
        \quad Elite ratio for crossover & 0.3 \\
        \quad Crossover rate & 1 \\
        \quad Mutation rate & 0.5 \\
        \quad CGES Softmax temperature & 1.0 \\
        \addlinespace

        \multicolumn{2}{@{}l}{\textbf{Large Language Model (LLM)}} \\
        \quad LLM model & Gemini 2.0 Flash \\
        \quad LLM temperature (generator \& reflector) & 1 \\
        \quad Max words for short-term reflection & 200 words \\
        \quad Max words for long-term reflection & 20 words \\
        \addlinespace

        \multicolumn{2}{@{}l}{\textbf{Trajectory Prediction}} \\
        \quad Num. prediction samples ($K$) & 20 \\
        \quad Observation length ($T_{\text{obs}}$) & 8 frames (3.2s) \\
        \quad Prediction length ($T_{\text{pred}}$) & 12 frames (4.8s) \\
        \bottomrule
    \end{tabular}
\end{table}

\subsection{Analysis of Exploration Ratio}
\label{appendix:ratio}

In our evolutionary framework, the selection of parents for crossover is governed by an exploration-exploitation trade-off. A portion of parents is chosen uniformly at random from all successful candidates (exploration), while the remainder is selected from the top-performing elite heuristics (exploitation). To determine the optimal balance for this process, we conducted a sensitivity analysis by varying the exploration ratio from 0.0 (pure exploitation) to 1.0 (pure exploration).

\begin{figure}[h!]
    \centering
    \includegraphics[width=0.8\linewidth]{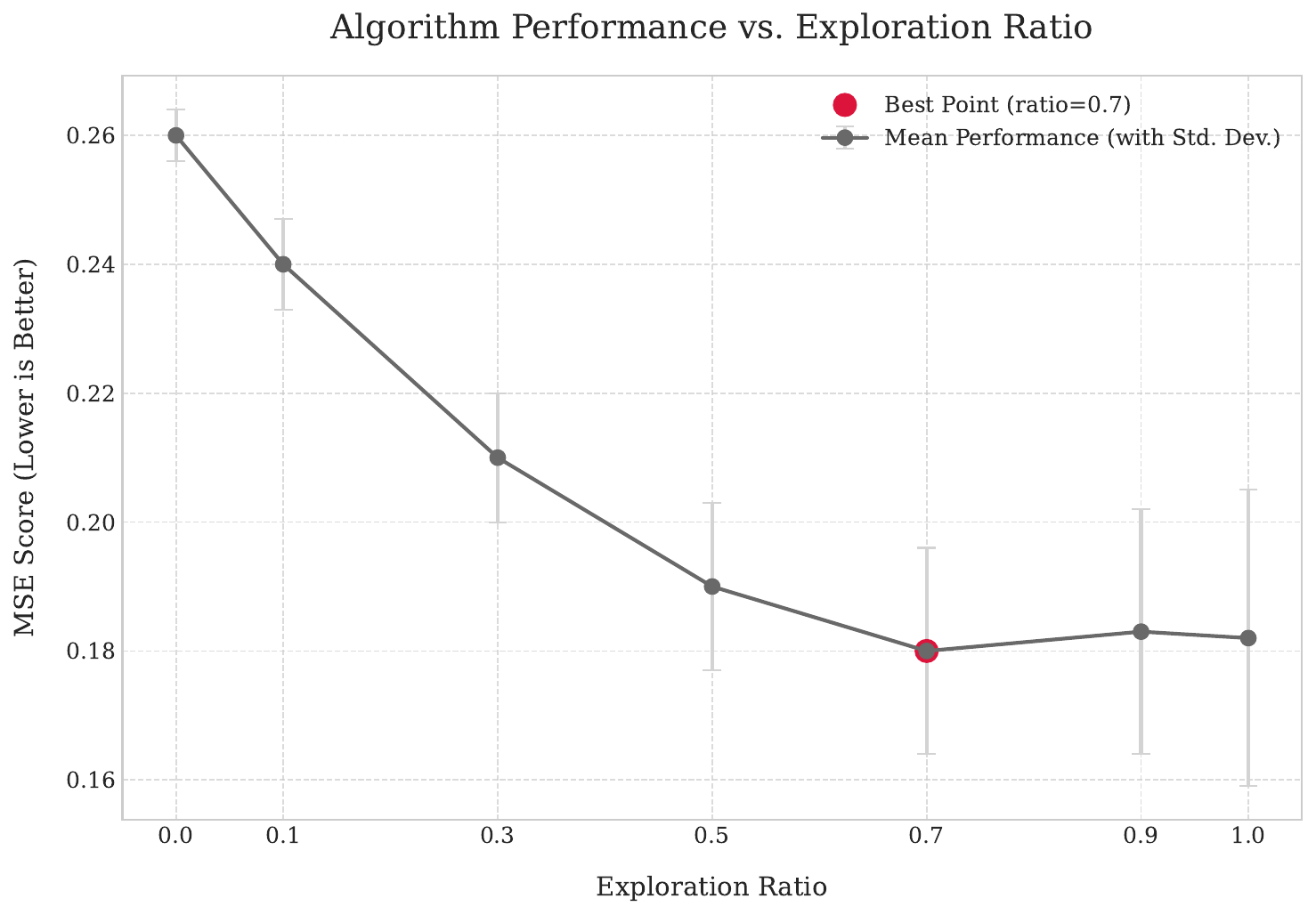}
    \caption{Performance sensitivity to the exploration ratio for crossover parent selection. The x-axis represents the proportion of parents selected uniformly at random (exploration), while the y-axis shows the resulting MSE score (lower is better). The optimal performance is achieved at an exploration ratio of 0.7, effectively balancing exploration and exploitation.}
    \label{fig:exploration_ratio}
\end{figure}

The results of this analysis are presented in Figure~\ref{fig:exploration_ratio}. The plot shows the final Mean Squared Error (MSE) score as a function of the exploration ratio. We observed that performance generally improves as the exploration ratio increases from 0.0, reaching its peak at a ratio of 0.7. Beyond this point, further increasing the exploration led to a slight degradation in the mean performance. While a higher exploration ratio could occasionally yield a superior result in a specific run due to greater randomness, it also introduced significant performance instability, as indicated by the larger standard deviation. Therefore, an exploration ratio of 0.7 was identified as the optimal setting, offering the best balance between high performance and reliable outcomes. This value was used for all experiments conducted in this paper.

\subsection{Detailed Resources}

\begin{figure}[htbp]
    \centering
    \includegraphics[width=0.8\linewidth]{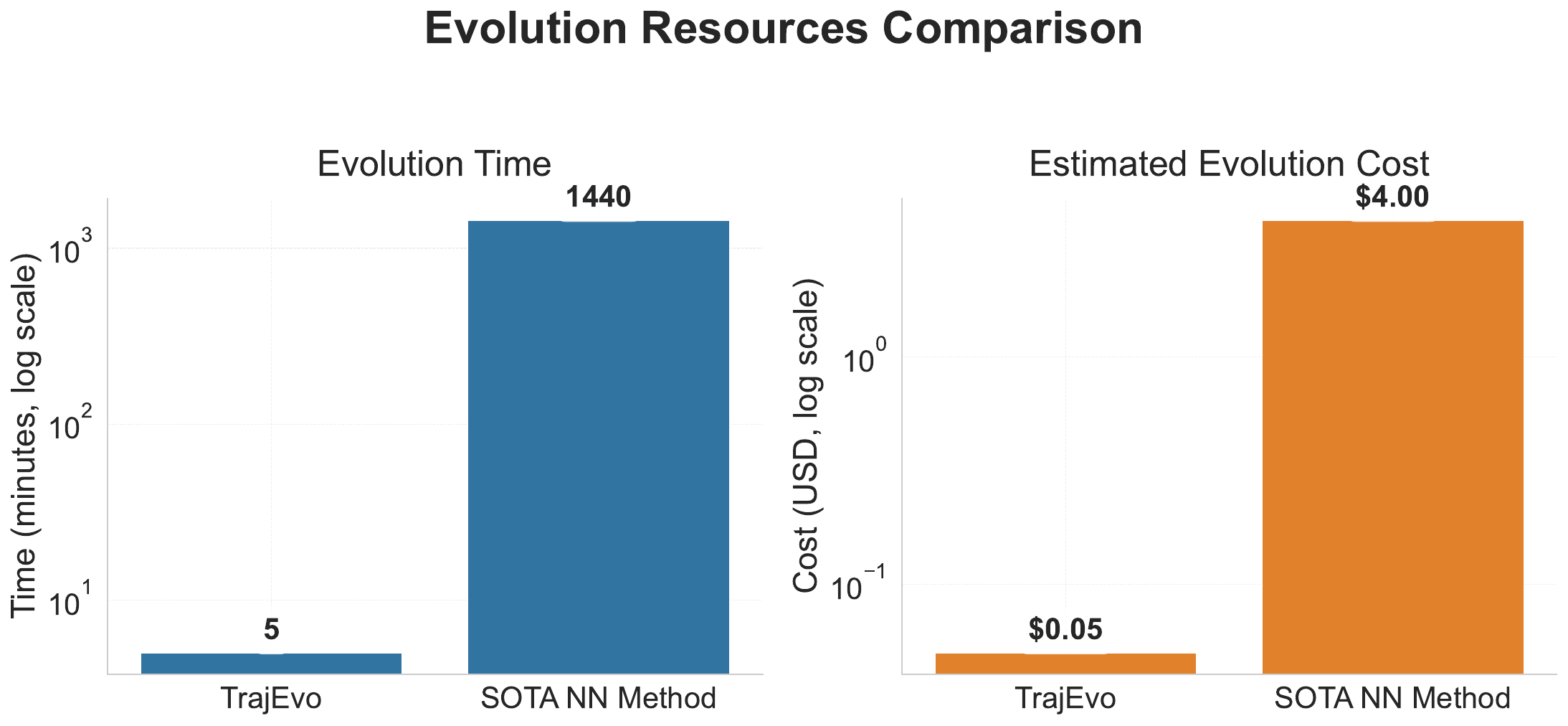}
    \caption{
        \textbf{Evolution Resources Comparison.}
        Compared to SOTA neural network methods, \our{} shows a significant advantage in both evolution time and cost (e.g., 5 minutes vs. 1 day, ~$0.05 vs. ~$4.00). Both metrics are displayed on a logarithmic scale.
    }
    \label{fig:evo_resources}
\end{figure}

\begin{figure}[htbp]
    \centering
    \includegraphics[width=0.8\linewidth]{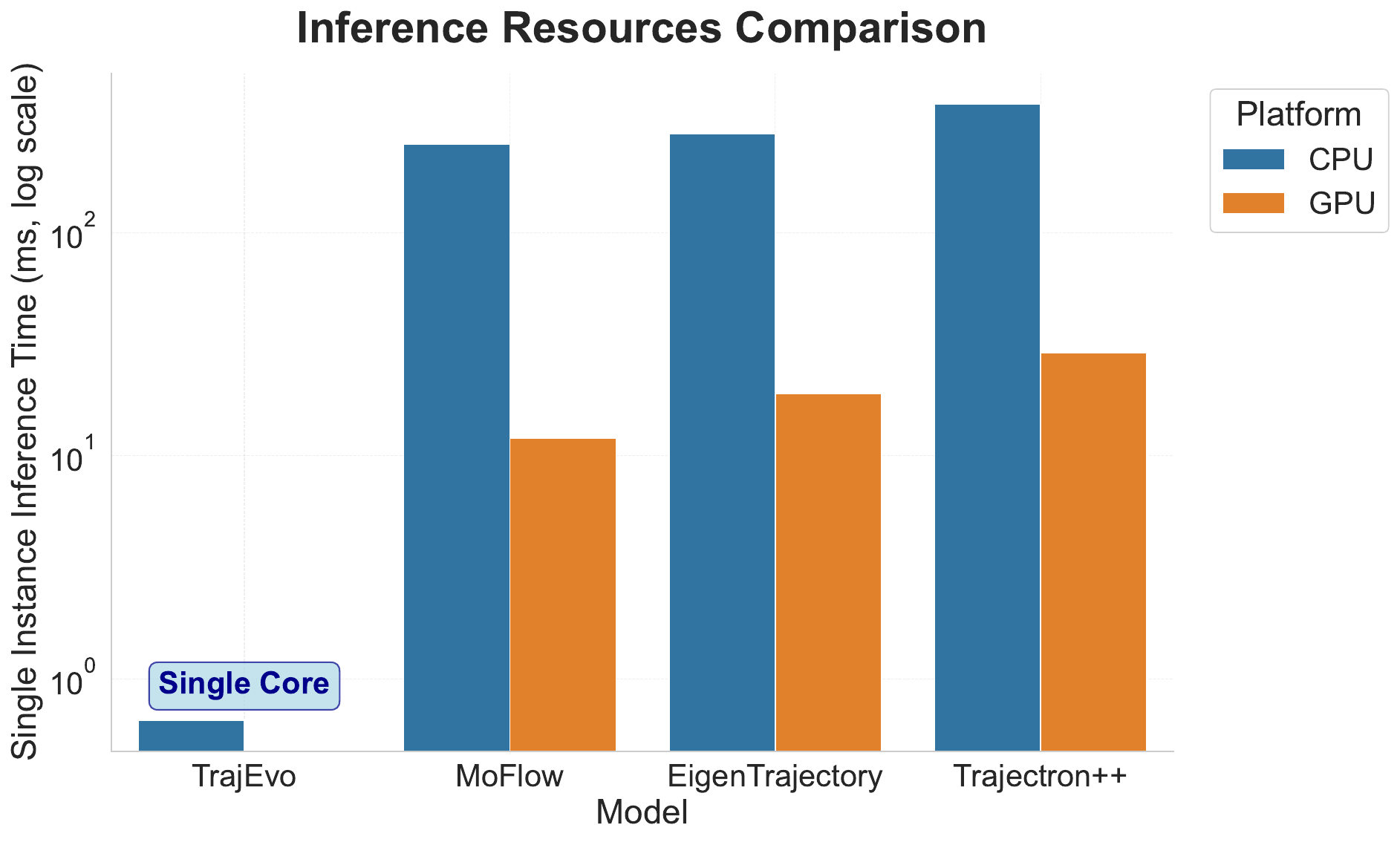}
    \caption{
        \textbf{Inference Resources Comparison.}
        \our{} achieves a fast 0.65ms inference time on a single CPU core, while baseline methods require more time even on a GPU. All times are displayed on a logarithmic scale.
    }
    \label{fig:inf_resources}
\end{figure}

\paragraph{Evolution and Inference Resources}
We conducted a comprehensive analysis of the computational resources required by \our{} and other baseline methods, covering both the one-time evolution cost and the per-instance inference cost. The results, detailed in \cref{fig:evo_resources} and \cref{fig:inf_resources}, show that \our{} is exceptionally efficient in both regards.

A single evolution run of \our{} takes approximately 5 minutes and costs about \$0.05. In contrast, a state-of-the-art (SOTA) neural method requires a full day of training, costing around \$4.00. This efficiency is highlighted in \cref{fig:evo_resources}. For real-time robotics applications, inference speed is critical. As shown in \cref{fig:inf_resources}, \our{}-generated heuristics require only 0.65ms on a single CPU core, while neural baselines are significantly slower, even on a GPU.

\subsection{Visualizing the Impact of CGES}
To complement the quantitative results presented in the main paper's ablation study (Table 4), we provide a visual illustration of the impact of Cross-Generation Elite Sampling (CGES) on the evolutionary process. Figure 9 plots the convergence of the objective value during the evolutionary search, comparing the full \our{} framework against an ablated version without CGES.

As the figure demonstrates, the inclusion of CGES allows the framework to escape local optima more effectively. The full \our{} framework (represented by the red line) consistently achieves a lower (better) objective value compared to the version without CGES (blue line). This visually confirms the data in Table 4 and underscores the importance of CGES for enhancing the quality and performance of the heuristics generated during evolution. The shaded areas represent the standard deviation across multiple runs, showing the consistency of this performance improvement.

\begin{figure}[h!]
    \centering
    \includegraphics[width=0.5\linewidth]{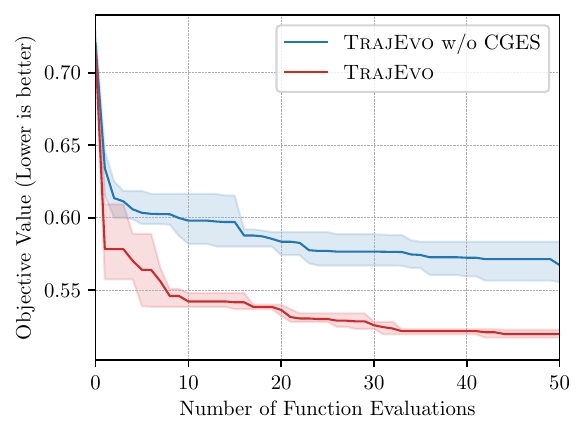}
    \caption{Visual comparison of the evolutionary process with and without CGES. The plot shows the objective value (lower is better) over the number of function evaluations, averaged over multiple runs. The \our{} framework with CGES consistently converges to better solutions than the ablated version, highlighting the effectiveness of CGES in improving the search process.}
    \label{fig:cges_ablation_appendix}
\end{figure}

\section{Limitations}

While \our{} introduces a promising paradigm for heuristic design in trajectory prediction, achieving a compelling balance of performance, efficiency, interpretability, and notably strong generalization (\cref{tab:cross-dataset-generalization}), we identify several limitations that also serve as important directions for future research:
\paragraph{In-Distribution Accuracy} 
Although \our{} significantly advances the state-of-the-art for heuristic methods and surpasses several deep learning baselines, the generated heuristics do not consistently achieve the absolute lowest error metrics on standard benchmarks compared to the most recent, highly specialized deep learning models when evaluated strictly \textit{in-distribution}. This likely reflects the inherent complexity trade-off; heuristics evolved for interpretability and speed may have a different expressivity limit compared to large neural networks. 
Future work could investigate techniques to further close this gap, potentially through more advanced evolutionary operators and heuristics integration into parts of simulation frameworks while preserving the core benefits.
\paragraph{Input Data Complexity} 
Our current evaluations focus on standard trajectory datasets using primarily positional history. Real-world robotic systems often have access to richer sensor data from which several features can be extracted, including agent types (pedestrians, vehicles), semantic maps (lanes, intersections), and perception outputs (detected obstacles, drivable space) -- for instance, given obstacle positions, we would expect \our{} to discover more likely trajectories that tend to avoid obstacles. \our{} currently does not leverage this complexity. Extending the framework to incorporate and reason about such inputs would represent a significant next step. This could enable the automatic discovery of heuristics that are more deeply context-aware and reactive to complex environmental factors.
\paragraph{Downstream Task Performance} 
We evaluate \our{} based on standard trajectory prediction metrics (minADE/minFDE). While these metrics often correlate to downstream task performance \citep{phong2023truly}, these may not always perfectly correlate with performance on downstream robotic tasks like navigation or planning. Further developing our framework to optimize heuristics directly for task-specific objectives within a closed loop (e.g., minimizing collisions or travel time in simulation) represents an interesting avenue for future works that could lead to more practically effective trajectory prediction.

\section{Prompts}
\label{appendix:prompts}

\subsection{Common prompts}
The prompt formats are given below for the main evolutionary framework of \our{}. These are based on ReEvo \citep{ye2024reevo}, but with modified prompts tailored specifically for the task of trajectory prediction heuristic design and to incorporate the unique mechanisms of \our{}. 

\renewcommand{\lstlistingname}{Prompt}
\crefname{lstlisting}{Prompt}{Prompts}

\begin{lstlisting}[caption={System prompt for generator LLM.},  label={lst: system prompt for generator LLM}, style=promptstyle_appendix]
You are an expert in the domain of prediction heuristics. Your task is to design heuristics that can effectively solve a prediction problem.
Your response outputs Python code and nothing else. Format your code as a Python code string: "```python ... ```".    
\end{lstlisting}

\begin{lstlisting}[caption={System prompt for reflector LLM.},  label={lst: system prompt for reflector LLM}, style=promptstyle_appendix]
You are an expert in the domain of prediction heuristics. Your task is to give hints to design better heuristics.
\end{lstlisting}

\begin{lstlisting}[caption={Task description.},  label={lst: task description}, style=promptstyle_appendix]
Write a {function_name} function for {problem_description}
{function_description}
\end{lstlisting}

\begin{lstlisting}[caption={User prompt for population initialization.},  label={lst: user prompt for population initialization}, style=promptstyle_appendix]
{task_description}

{seed_function}

Refer to the format of a trivial design above. Be very creative and give `{func_name}_v2`. Output code only and enclose your code with Python code block: ```python ... ```.

{initial_long-term_reflection}
\end{lstlisting}

\begin{lstlisting}[caption={User prompt for crossover.},  label={lst: user prompt for crossover}, style=promptstyle_appendix]
{task_description}

[Worse code]
{function_signature0}
{worse_code}

[Better code]
{function_signature1}
{better_code}

[Reflection]
{short_term_reflection}

[Improved code]
Please write an improved function `{function_name}_v2`, according to the reflection. Output code only and enclose your code with Python code block: ```python ... ```.
\end{lstlisting}
\paragraph{Integration of Statistics Feedback Loop}
The prompts for reflection and mutation are designed to leverage \our{}'s Statistics Feedback Loop. \\
Specifically, the short-term reflection prompt (\cref{lst:user-prompt-short-termreflection}) and the elitist mutation prompt (\cref{lst: user prompt for elitist mutation}) explicitly require the LLM to consider "trajectory statistics" or "Code Results Analysis" when generating reflections or new heuristic code. This allows the LLM to make data-driven decisions based on the empirical performance of different heuristic strategies.

\begin{lstlisting}[caption={User prompt for short-term reflection.},  label={lst:user-prompt-short-termreflection}, style=promptstyle_appendix]
Below are two {func_name} functions for {problem_desc}
{func_desc}

You are provided with two code versions below, where the second version performs better than the first one.

[Worse code]
{worse_code}

[Worse code results analysis]
{stats_info_worse}

[Better code]
{better_code}

[Better code results analysis]

{stats_info_better}

Respond with some hints for designing better heuristics, based on the two code versions and the trajectory statistics. Be concise. Use a maximum of 200 words.
\end{lstlisting}

\begin{lstlisting}[caption={User prompt for long-term reflection.},  label={lst: user prompt for long-term reflection}, style=promptstyle_appendix]
Below are two {function_name} functions for {problem_description}
{function_description}

You are provided with two code versions below, where the second version performs better than the first one.

[Worse code]
{worse_code}

[Better code]
{better_code}

You respond with some hints for designing better heuristics, based on the two code versions and using less than 20 words.
\end{lstlisting}

\begin{lstlisting}[caption={User prompt for elitist mutation.},  label={lst: user prompt for elitist mutation}, style=promptstyle_appendix]
{user_generator}

[Prior reflection]
{reflection}

[Code]
{func_signature1}
{elitist_code}

[Code Results Analysis]
{stats_info_elitist}

[Improved code]
Please write a mutated function `{func_name}_v2`, according to the reflection. Output code only and enclose your code with Python code block: ```python ... ```. 

Please generate mutation versions that are significantly different from the base code to increase exploration diversity.
\end{lstlisting}

\subsection{Trajectory Prediction-specific Prompts}

\paragraph{Domain Specialization} All prompts are contextualized for the domain of trajectory prediction heuristics. For example, the system prompts are deeply contextualized through specific prompts detailing the trajectory prediction problem:

\begin{lstlisting}[caption={Function Signature},  label={lst: function signature}, style=promptstyle_appendix]]
def predict_trajectory{version}(trajectory: np.ndarray) -> np.ndarray:
\end{lstlisting}

\begin{lstlisting}[caption={Function Description},  label={lst: function description}, style=promptstyle_appendix]]
The predict_trajectory function takes as input the current trajectory (8 frames) and generates 20 possible future trajectories for the next 12 frames. It has only one parameter: the past trajectory array.
The output is a numpy array of shape [20, num_agents, 12, 2] containing all 20 trajectories.
Note that we are interesting in obtaining at least one good trajectory, not necessarily 20.
Thus, diversifying a little bit is good.
Note that the heuristic should be generalizable to new distributions.
\end{lstlisting}

\begin{lstlisting}[caption={Seed Function}, label={lst: seed function}, style=promptstyle_appendix]]
def predict_trajectory(trajectory):
    """Generate 20 possible future trajectories
    Args:
        - trajectory [num_agents, traj_length, 2]: here the traj_length is 8;
    Returns:
        - 20 diverse trajectories [20, num_agents, 12, 2]
    """
    all_trajectories = []
    for _ in range(20):
        current_pos = trajectory[:, -1, :]
        velocity = trajectory[:, -1, :] - trajectory[:, -2, :] # only use the last two frames
        predictions = []
        for t in range(1, 12+1): # 12 future frames
            current_pos = current_pos + velocity * 1 # dt
            predictions.append(current_pos.copy())
        pred_trajectory = np.stack(predictions, axis=1)
        all_trajectories.append(pred_trajectory)
    all_trajectories = np.stack(all_trajectories, axis=0)
    return all_trajectories
\end{lstlisting}

\begin{lstlisting}[caption={External Knowledge},  label={lst: external knowledge}, style=promptstyle_appendix]]
# External Knowledge for Pedestrian Trajectory Prediction

## Task Definition
- We are using the ETH/UCY dataset for this task (human trajectory prediction)
- Input: Past 8 frames of pedestrian positions
- Output: Future 12 frames of pedestrian positions
- Variable number of pedestrians per scene
\end{lstlisting}

\section{\our{} Output}

\subsection{Generated Heuristics}

\renewcommand{\lstlistingname}{Heuristic}
\crefname{lstlisting}{Heuristic}{Heuristics}

\begin{lstlisting}[caption={The best \our{}-generated heuristic for Zara 1.},  label={lst:heuristic-eth}, language=Python, style=heuristicstyle]
import numpy as np

def predict_trajectory(trajectory: np.ndarray) -> np.ndarray:
    """Generate 20 possible future trajectories with enhanced diversification and adaptive strategies.

    Args:
        trajectory (np.ndarray): [num_agents, traj_length, 2] where traj_length is 8.

    Returns:
        np.ndarray: 20 diverse trajectories [20, num_agents, 12, 2].
    """
    num_agents = trajectory.shape[0]
    all_trajectories = []
    history_len = trajectory.shape[1]

    for i in range(20):
        current_pos = trajectory[:, -1, :]

        # Option 1: Dominant strategy - Average velocity with adaptive noise, rotation, and parameter variation
        if i < 14:  # Increased to 14, best performing strategy
            velocity = np.zeros_like(current_pos)
            weights_sum = 0.0
            decay_rate = np.random.uniform(0.1, 0.3)  # Adaptive decay rate
            for k in range(min(history_len - 1, 5)):
                weight = np.exp(-decay_rate * k)
                velocity += weight * (trajectory[:, -1 - k, :] - trajectory[:, -2 - k, :])
                weights_sum += weight
            velocity /= (weights_sum + 1e-8)

            avg_speed = np.mean(np.linalg.norm(velocity, axis=1))
            noise_scale = 0.012 + avg_speed * 0.008
            noise = np.random.normal(0, noise_scale, size=(num_agents, 12, 2))

            angle = np.random.uniform(-0.05, 0.05)
            rotation_matrix = np.array([[np.cos(angle), -np.sin(angle)],
                                        [np.sin(angle), np.cos(angle)]])
            velocity = velocity @ rotation_matrix

            momentum = 0.0
            jerk_factor = 0.0
            damping = 0.0

            # Parameter Variation
            if i % 6 == 0:
                noise_scale *= np.random.uniform(0.9, 1.1)  # Fine-tuned noise scale variation
                noise = np.random.normal(0, noise_scale, size=(num_agents, 12, 2))
            elif i % 6 == 1:
                angle_scale = 0.06 + avg_speed * 0.02
                angle = np.random.uniform(-angle_scale * np.random.uniform(0.8, 1.2), angle_scale * np.random.uniform(0.8, 1.2))
                rotation_matrix = np.array([[np.cos(angle), -np.sin(angle)],
                                            [np.sin(angle), np.cos(angle)]])
                velocity = velocity @ rotation_matrix
            elif i % 6 == 2:
                momentum = np.random.uniform(0.06, 0.14)  # Vary momentum
                velocity = momentum * velocity + (1 - momentum) * (trajectory[:, -1, :] - trajectory[:, -2, :])
            elif i % 6 == 3:  # Add jerk
                jerk_factor = np.random.uniform(0.0025, 0.0065)
                if history_len > 2:
                    jerk = (trajectory[:, -1, :] - 2 * trajectory[:, -2, :] + trajectory[:, -3, :])
                else:
                    jerk = np.zeros_like(velocity)
                velocity += jerk_factor * jerk
            elif i % 6 == 4: # Damping
                damping = np.random.uniform(0.006, 0.019)
                velocity = velocity * (1 - damping)
            else: # Adaptive Noise Scale
                noise_scale = 0.01 + avg_speed * np.random.uniform(0.006, 0.014)
                noise = np.random.normal(0, noise_scale, size=(num_agents, 12, 2))


            predictions = []
            for t in range(1, 13):
                current_pos = current_pos + velocity + noise[:, t-1, :] / (t**0.4)
                predictions.append(current_pos.copy())
            pred_trajectory = np.stack(predictions, axis=1)

        # Option 2: Velocity rotation with adaptive angle
        elif i < 17:  # Increased to 17.
            velocity = trajectory[:, -1, :] - trajectory[:, -2, :]
            avg_speed = np.mean(np.linalg.norm(velocity, axis=1))
            angle_scale = 0.13 + avg_speed * 0.05  # adaptive angle

            angle = np.random.uniform(-angle_scale, angle_scale)  # adaptive range
            rotation_matrix = np.array([[np.cos(angle), -np.sin(angle)],
                                        [np.sin(angle), np.cos(angle)]])
            velocity = velocity @ rotation_matrix

            noise_scale = 0.007 + avg_speed * 0.004
            noise = np.random.normal(0, noise_scale, size=(num_agents, 12, 2))

            predictions = []
            for t in range(1, 13):
                current_pos = current_pos + velocity + noise[:, t-1, :] / (t**0.5)
                predictions.append(current_pos.copy())
            pred_trajectory = np.stack(predictions, axis=1)

        # Option 3: Memory-based approach (repeating last velocity) + Enhanced Collision Avoidance
        elif i < 19:  # Increased to 19
            velocity = trajectory[:, -1, :] - trajectory[:, -2, :]
            # Enhanced smoothing with more velocity history
            if history_len > 3:
                velocity = 0.55 * velocity + 0.3 * (trajectory[:, -2, :] - trajectory[:, -3, :]) + 0.15 * (trajectory[:, -3, :] - trajectory[:, -4, :])
            elif history_len > 2:
                velocity = 0.65 * velocity + 0.35 * (trajectory[:, -2, :] - trajectory[:, -3, :])
            else:
                velocity = velocity # do nothing

            avg_speed = np.mean(np.linalg.norm(velocity, axis=1))

            # Adaptive Laplacian noise
            noise_scale = 0.005 + avg_speed * 0.0015
            noise = np.random.laplace(0, noise_scale, size=(num_agents, 2))
            velocity = velocity + noise

            # Enhanced collision avoidance
            repulsion_strength = 0.0011 # Adjusted repulsion strength
            predictions = []
            temp_pos = current_pos.copy()

            # Store predicted positions for efficient collision calculation at each timestep
            future_positions = [temp_pos.copy()]  # Start with current position
            for t in range(1, 13):
                net_repulsions = np.zeros_like(temp_pos)
                for agent_idx in range(num_agents):
                    for other_idx in range(num_agents):
                        if agent_idx != other_idx:
                            direction = temp_pos[agent_idx] - temp_pos[other_idx]
                            distance = np.linalg.norm(direction)
                            if distance < 1.05:  # Adjusted interaction threshold
                                repulsion = (direction / (distance + 1e-6)) * repulsion_strength * np.exp(-distance)  # distance-based decay
                                net_repulsions[agent_idx] += repulsion

                velocity = 0.9 * velocity + 0.1 * net_repulsions # Damping the change in velocity
                temp_pos = temp_pos + velocity
                future_positions.append(temp_pos.copy()) # Store for future repulsion calculations
                predictions.append(temp_pos.copy())

            pred_trajectory = np.stack(predictions, axis=1)

        # Option 4: Linear prediction with adaptive damping and larger noise.
        else:
            velocity = trajectory[:, -1, :] - trajectory[:, -2, :]
            damping = np.random.uniform(0.017, 0.038)  # damping factor

            noise_scale = 0.028
            noise = np.random.normal(0, noise_scale, size=(num_agents, 12, 2))

            predictions = []
            for t in range(1, 13):
                velocity = velocity * (1-damping) + noise[:, t-1, :] / (t**0.4)  # damping
                current_pos = current_pos + velocity
                predictions.append(current_pos.copy())
            pred_trajectory = np.stack(predictions, axis=1)

        all_trajectories.append(pred_trajectory)

    all_trajectories = np.stack(all_trajectories, axis=0)
    return all_trajectories
\end{lstlisting}

\subsection{Reflections}

\renewcommand{\lstlistingname}{Output}
\crefname{lstlisting}{Output}{Outputs}

\begin{lstlisting}[caption={Long-term reasoning output},  label={lst:reasoning}, style=promptstyle_appendix]]
Based on comparative analysis, prioritize these heuristics:

1.  **Hierarchical Stochasticity:** Sample trajectory-level parameters (speed scale, movement pattern) *once* per trajectory. Then, apply agent-specific stochastic variations within those constraints. Introduce `global_randomness` sampled *once* per trajectory to couple different parameters.
2.  **Adaptive Movement Primitives:** Condition movement model probabilities (stop, turn, straight, lane change, obstacle avoidance) on agent state (speed, acceleration, past turning behavior, context). Consider longer history windows.
3.  **Refine Noise & Parameters:** Finetune noise scales and apply dampening. Experiment with learnable parameters and wider ranges. Directly manipulate velocity and acceleration stochastically for smoother transitions.
4.  **Contextual Interactions:** Enhance social force models, considering intentions, agent types, and environment.
5.  **Guaranteed Diversity:** Ensure movement probabilities sum to 1.
6.  **Post Processing:** Apply smoothing and collision avoidance.
7. **Intentions:** Incorporate high level intentions such as "going to an area."
\end{lstlisting}

\cref{lst:reasoning}  shows an example of long-term reasoning output for the model, based on the comparative analysis. 
\our{} discovers several interesting heuristics for trajectory forecasting, such as applying diverse noise factors, social force models, diversity, and modeling intentions to model possible future trajectories.

\cref{lst: interactions} shows some more outputs of \our{} from various runs, which discovers some interesting helper functions that model interactions such as stochastic, social force, and diversity.

\begin{lstlisting}[caption={Selected \our{} Interactions}, label={lst: interactions}, language=Python, style=heuristicstyle]]
################### 
# Model with Noise
################### 

def acceleration_model_with_noise(trajectory, noise_level_base=0.05, prediction_steps=12):
    velocity = trajectory[:, -1, :] - trajectory[:, -2, :]
    acceleration = velocity - (trajectory[:, -2, :] - trajectory[:, -3, :])
    current_pos = trajectory[:, -1, :].copy()
    current_velocity = velocity.copy()
    predictions = []
    for i in range(prediction_steps):
        noise_level = noise_level_base * (i + 1)
        current_velocity = current_velocity + acceleration + np.random.normal(0, noise_level, size=current_velocity.shape)
        current_pos = current_pos + current_velocity
        predictions.append(current_pos.copy())
    return np.stack(predictions, axis=1)

    
################### 
# Social Force
################### 

def constant_velocity_with_repulsion(trajectory, get_nearby_agents, repulsion_strength=0.05, num_steps=12):
    velocity = trajectory[:, -1, :] - trajectory[:, -2, :]
    current_pos = trajectory[:, -1, :].copy()
    repulsion_force = np.zeros_like(current_pos)
    num_agents = trajectory.shape[0]

    for _ in range(num_steps):
        for agent_index in range(num_agents):
            nearby_agents = get_nearby_agents(agent_index, current_pos)
            for neighbor_index in nearby_agents:
                direction = current_pos[agent_index] - current_pos[neighbor_index]
                distance = np.linalg.norm(direction)
                if distance > 0:
                    repulsion_force[agent_index] += (direction / (distance**2 + 0.001)) * repulsion_strength
    return repulsion_force

    
################### 
# Diversity
################### 

def simple_random_walk(trajectory, noise_level_base=0.2, prediction_steps=12):
    current_pos = trajectory[:, -1, :].copy()
    predictions = []
    for _ in range(prediction_steps):
        noise_level = noise_level_base * (_ + 1)
        current_pos = current_pos + np.random.normal(0, noise_level, size=current_pos.shape)
        predictions.append(current_pos.copy())
    return np.stack(predictions, axis=1)
\end{lstlisting}

\end{document}